\documentclass{article}

\PassOptionsToPackage{numbers}{natbib}

\usepackage[preprint]{neurips_2019}

\usepackage{float}
\usepackage[utf8]{inputenc} 
\usepackage[T1]{fontenc}    
\usepackage{microtype}      
\usepackage{color,graphicx}
\usepackage{hyperref}       
\usepackage{url}            
\usepackage{booktabs}       
\usepackage{amsfonts}       
\usepackage{nicefrac}       
\usepackage{amsmath}
\usepackage{subcaption}
\usepackage{wrapfig}
\usepackage{todonotes}
\usepackage{multicol}
\usepackage{multirow}
\usepackage{algorithm}
\usepackage[noend]{algpseudocode}
\usepackage{amssymb}
\usepackage{amsmath}
\usepackage{amsthm}
\usepackage{bm}
\usepackage{pifont}
\usepackage{color}

\newboolean{include-notes}
\setboolean{include-notes}{false}
\newcommand{\dmkr}[1]{\ifthenelse{\boolean{include-notes}}
 {{\color{purple} D: #1}}{}}
\newcommand{\rs}[1]{\ifthenelse{\boolean{include-notes}}
 {{\color{blue} R: #1}}{}}
\newcommand{\herke}[1]{\ifthenelse{\boolean{include-notes}}
 {{\color{red} H: #1}}{}}
 \newcommand{\herkenew}[1]{\ifthenelse{\boolean{include-notes}}
 {{\color{violet} H: #1}}{}}
\newcommand{\lawrence}[1]{\ifthenelse{\boolean{include-notes}}
 {{\color{green} L: #1}}{}}

{\begin{list}{$\bullet$}{%
    \setlength{\topsep}{0in}
    \setlength{\partopsep}{0in}
    \setlength{\itemsep}{0in}
    \setlength{\parsep}{0in}
    \setlength{\leftmargin}{3.5em}
    \setlength{\rightmargin}{0in}
    \setlength{\itemindent}{-.1in}
}
}%
{\end{list}
}

\DeclareMathOperator*{\argmax}{argmax}
\newcommand{\expect}[2]{\mathbb{E}_{#1}\left[#2\right]}

\DeclareMathOperator{\Dirichlet}{Dir}
\DeclareMathOperator{\Beta}{Beta}

\DeclareMathOperator{\true}{true}

\DeclareMathOperator{\supp}{supp}

\newcommand{\fe}[1]{\mathcal{F}^{#1}}
\newcommand{\tin}[1]{\widetilde{\theta}^{#1}}
\newcommand{\tout}{\theta^*}
\newcommand{\posterior}{p(\tout \mid \tin{1}, \tin{2})}
\newcommand{\tindep}{\Theta^{\text{indep pts}}}

\definecolor{mygreen}{RGB}{34,139,34}
\newcommand{\yescheck}{\textcolor{mygreen}{\ding{51}}}
\newcommand{\nocross}{\textcolor{red}{\ding{55}}}
\newcommand{\kindof}{{$\approx$}}

\let\emptyset\varnothing

\newtheorem{theorem}{Theorem}
\newtheorem{desideratum}{Desideratum}
\newtheorem{corollary}{Corollary}[theorem]

\newcommand{\prg}[1]{\noindent\textbf{#1. }} 

\title{Combining reward information from multiple sources}

\author{%
  Dmitrii Krasheninnikov \thanks{Work done at UC Berkeley}  \hspace{1pt} \thanks{Correspondence to \texttt{dmitrii.krasheninnikov@sony.com, rohinmshah@berkeley.edu}} \\
  University of Amsterdam
  \And
  Rohin Shah \footnotemark[2] \\
  UC Berkeley
  \And
  Herke van Hoof \\
  University of Amsterdam
}

\begin{document}

\maketitle

\begin{abstract}
Given two sources of evidence about a latent variable, one can combine the information from both by multiplying the likelihoods of each piece of evidence. However, when one or both of the observation models are \emph{misspecified}, the distributions will conflict. We study this problem in the setting with two conflicting reward functions learned from different sources. In such a setting, we would like to retreat to a broader distribution over reward functions, in order to mitigate the effects of misspecification. We assume that an agent will maximize expected reward given this distribution over reward functions, and identify four desiderata for this setting. We propose a novel algorithm, Multitask Inverse Reward Design (MIRD), and compare it to a range of simple baselines. While all methods must trade off between conservatism and informativeness, through a combination of theory and empirical results on a toy environment, we find that MIRD and its variant MIRD-IF strike a good balance between the two.                                 
\end{abstract}

\section{Introduction}

\herke{General comment: for me the notion/notation of p(theta) as (posterior) probability distribution is a bit weird. It is not exactly a posterior. Is it just 'whatever vector can we put into AUP to make it behave as we want' without intrinsic meaning? Maybe it is better to think of it as 'weight' of each particular reward hypothesis?}

\rs{Added a comment to related work, and I agree with Lawrence's comment about there being more future work, but both of those comments are not that important and it's fine if you ignore both. I also looked at all the other comments you made and either addressed + deleted them, or gave my opinion on them which you can take as you will.}

While deep reinforcement learning (RL) has led to considerable success when \emph{given} an accurate reward function~\cite{alphago, OpenAI_dota}, \emph{specifying} a reward function that captures human preferences in real-world tasks is challenging~\citep{amodei2016concrete, christiano2017deep}. Value learning seeks to sidestep this difficulty by learning rewards from various types of data, such as natural language~\citep{NaturalLanguagePreferenceElicitation}, demonstrations~\citep{ziebart2010modeling, fu2017learning}, comparisons~\citep{christiano2017deep}, ratings~\citep{daniel2014active}, human reinforcement~\citep{COACH}, proxy rewards~\citep{hadfield2017inverse}, the state of the world~\cite{rlsp}, etc. Many approaches to value learning aim to create a distribution over reward functions~\citep{hadfield2016cooperative,hadfield2017inverse, ramachandran2007bayesian, dorsa2017active}. 

\begin{wrapfigure}[12]{h}{0.48\textwidth}
    \centering
    \includegraphics[width=0.48\textwidth]{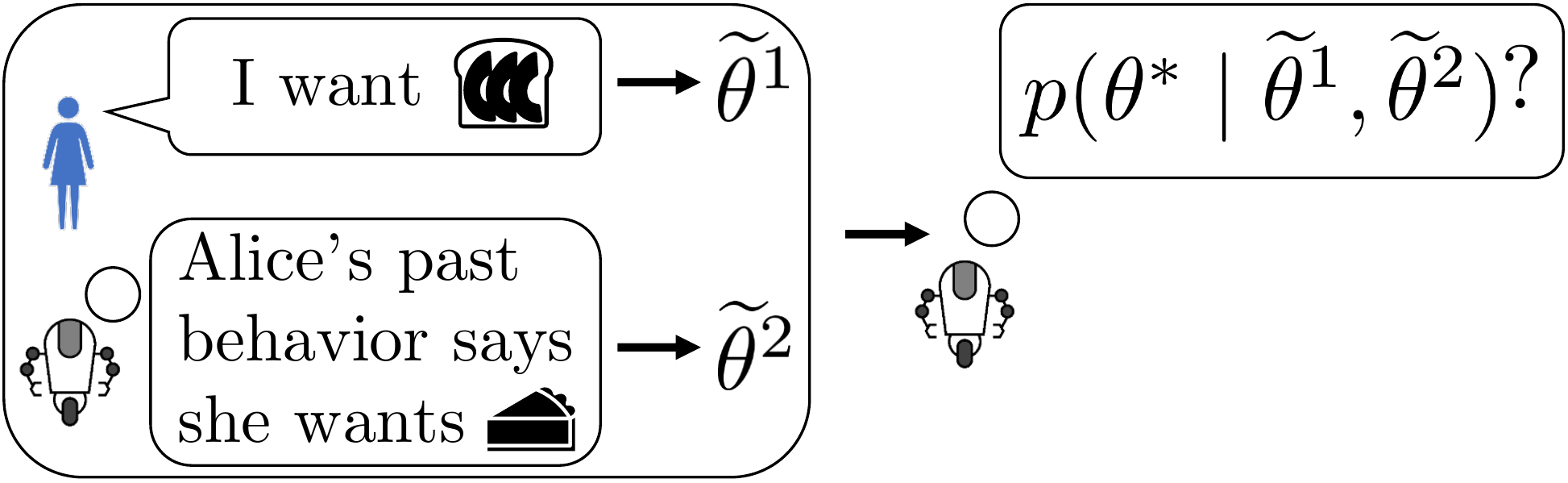}
\caption{Alice says she wants avocado toast, but her past actions imply she prefers cake to toast. What should the agent believe about the true reward $\tout$?}
\label{fig:front}
\end{wrapfigure}

A natural idea is to \emph{combine} the information from these diverse sources. However, the sources may conflict. For example, in Figure~\ref{fig:front}, Alice might say that she wants the healthy avocado toast, but empirically she almost always chooses to eat the tasty cake. A reward learning method based on natural language may confidently infer that Alice would prefer the toast, whereas a method based on revealed preferences 
might confidently infer that she wants cake. Presumably one of the algorithms made a poor assumption: perhaps the natural language algorithm failed to model that Alice only said that she wants avocado toast to impress her friends. Alternatively, maybe Alice wants to start eating healthy, but chooses the cake in moments of weakness, and the revealed preferences algorithm failed to model this bias. While it is unclear what the ``true'' reward is, we do know that at least one of the methods is \emph{confidently} wrong. We would like to defer on the question of what the ``true'' reward is, and instead retreat to a position of uncertainty. Formally, we want a distribution over the parameters of the true reward function $\tout$ given the inferred rewards $\tin{1}$ and $\tin{2}$. 

We could apply Bayes Rule: $p(\tout \mid \tin{1}, \tin{2}) \propto p(\tout) p(\tin{1} \mid \tout) p(\tin{2} \mid \tout)$ under the assumption that $\tin{1}$ and $\tin{2}$ are conditionally independent given $\tout$. However, when rewards conflict due to misspecification, for \emph{every} $\tout$ one of the likelihoods will be very low, and the posterior will be low everywhere, likely leading to garbage after normalization~\citep{frazier2017model}. While it is possible to mitigate this by making the assumptions more realistic, e.g. by modeling human biases~\citep{shah2019feasibility,RiskIRL,IgnorantInconsistentAgents}, it is very hard to get a perfect model~\citep{armstrong2018occam, steinhardt_evans_2017}. So, we aim for a method that is robust to misspecification.



What makes for a good reward combination method? To answer this question, we must know how the resulting distribution will be used. As in \citet{hadfield2016cooperative}, we assume that an agent maximizes expected reward, \emph{given that the agent can gather more information about the reward}. In this framework, an agent must balance three concerns: actively learning about the reward~\cite{dorsa2017active, mindermann2018active}, preserving its ability to pursue the true reward in the future when it knows more~\citep{turner2019conservative,krakovna2018penalizing}, and acting to gain immediate reward. Each concern suggests different desiderata for a reward combination method, which we formalize for rewards functions that are linear in features of the state.

There are several reasonable methods for combining reward functions based on the intuition that we should put weight on any parameter values ``in between'' the parameters of the given reward functions. These \emph{reward-space} methods operate in the space of parameters of the reward. 
However, this ignores any details about the environment and its features.\lawrence{The next sentence is somewhat confusing, even though I know what it is.} We introduce a method called Multi-task Inverse Reward Design (MIRD) that leverages the environment to infer which reward functions are compatible with the behaviors ``in between'' the behaviors incentivized by the two input rewards, making it a \emph{behavior-space} method. 
Through a combination of theory and empirical results on a simple environment, we evaluate how well the proposed methods meet our desiderata, and conclude that MIRD is a good option when the reward distribution must be used to act, while its variant MIRD-IF is better for increased robustness and a higher chance of support on the true reward.

\herke{Maybe give a one-sentence description of MIRD-IF?}

\lawrence{Maybe include a contributions section.}

\section{Background}\label{chp:background}
\textbf{Markov Decision Process (MDP).}
A MDP $\mathcal M$ is a tuple $\mathcal M = \langle \mathcal S, \mathcal A, \mathcal T, r, H \rangle$, where $\mathcal S$ is the set of states, $\mathcal A$ is the set of actions, $\mathcal T: \mathcal  S \times \mathcal  A \times \mathcal S \mapsto [0,1]$ is the transition probability function, $r : \mathcal  S \mapsto \mathbb{R}$ is the reward function, and $H \in \mathbb{Z}_{+}$ is the finite planning horizon. We consider MDPs where the reward is linear in features, and does not depend on action: $r(s ; \theta) = \theta^T f(s)$, where $\theta$ are the parameters defining the reward function and $f$ computes features of a given state. A policy $\pi : \mathcal{S} \mapsto \mathcal{A}$ specifies how to act in the MDP. A trajectory $\tau : (\mathcal{S} \times \mathcal{A})^*$ 
is a sequence of states and actions, where the actions are sampled from a policy $\pi$ and the states are sampled from the transition function $\mathcal{T}$. We abuse notation and write $f(\tau)$ to denote $\sum_{s \in \tau} f(s)$. The \emph{feature expectations} (FE) of policy $\pi$ are the expected feature counts when acting according to $\pi$: $\mathcal{F}^{\pi} = \mathbb{E}_{\tau \sim \pi}[f(\tau)]$. 
For conciseness we denote the feature expectations arising from optimizing reward $\theta$ as $\fe{\theta} = \fe{\pi_\theta}$. 



\textbf{Inverse Reinforcement Learning (IRL).}
In IRL the goal is to infer the reward parameters $\theta$ given a MDP without reward $\mathcal M \backslash r$ and expert demonstrations $D= \{ \tau_1, ..., \tau_m    \}$. 


\textbf{Maximum causal entropy inverse RL (MCEIRL).}\label{sec:mceirl}
As human demonstrations are rarely optimal, \citet{ziebart2010modeling} models the expert as a noisily rational agent that acts close to randomly when the difference in the actions' expected returns is small, but nearly always chooses the best action when it leads to a much higher expected return. 
Formally, MCEIRL models the expert as using soft value iteration to optimize its reward function: $\pi(a\mid s) = \exp[k (Q_t(s,a) - V(s))]$, where $k$ is the "rationality" parameter, and $V(s)~=~\frac{1}{k}\ln\sum_a \exp[kQ(s,a)]$ plays the role of a normalizing constant. The state-action value function $Q$ is computed as $\label{eq:soft-bellman}Q_t(s,a) = r(s) + \sum_{s'} \mathcal T(s' \mid s, a) V_{t+1}(s').$

\textbf{Reward learning by simulating the past (RLSP).}~\label{sec:rlsp}
\citet{rlsp} note that environments where humans have acted are already optimized for human preferences, and so contain reward information. The 
RLSP algorithm considers the current state $s_0$ as the final state in a human trajectory $\tau = (s_{-H}, a_{-H}, ..., s_0, a_0)$ generated by the human with preferences $\theta$. The RLSP likelihood $p(\theta \mid s_0)$ is obtained by marginalizing out $(s_{-H}, a_{-H}, \dots s_{-1}, a_{-1}, a_0)$, since only $s_0$ is actually observed.


\textbf{Inverse Reward Design (IRD).}~\label{sec:ird}
Inverse reward design (IRD) notes that since reward designers often produce rewards by checking what the reward does in the training environment using an iterative trial-and-error process, the final reward they produce is likely to produce good behavior \emph{in the environments that they used during the reward design process}. So, $\tin{}$ need not be identical to $\tout$, 
but merely provides evidence about it. Formally, IRD assumes that the probability of specifying a given $\tin{}$ is proportional to the exponent of the expected true return in the training environment: $p(\tin{} \mid \theta^*) \propto \exp ( {\theta^*}^\top \fe{\tin{}} )$. IRD inverts this generative model of \textit{reward design} to sample from the posterior of the true reward: $p( \theta^* \mid \tin{}) \propto p(\tin{} \mid \theta^*) p(\theta^*)$.

\section{Uses of the reward distribution}\label{chp:applications}

So far, we have argued that we would like to combine the information from two different sources into a single distribution over reward functions. But what makes such a distribution good? We take a pragmatic approach: we identify downstream tasks for such a distribution, find desiderata for a reward distribution for these tasks, and use these desiderata to guide our search for methods.

\subsection{Optimizing the reward}

\herke{bit redundant with earlier mention in intro} We follow the framework of \citet{hadfield2016cooperative} in which an agent maximizes expected return under reward uncertainty. 
In this setting, maximizing the return causes the agent to use its actions both to obtain (expected) reward, and to gather information about the true reward in order to better obtain reward in the future~\citep{hadfield2016cooperative, woodward2019learning}. We might expect that an agent could first learn the reward and then act; however, this is unrealistic: we would be unhappy with a household robot that had to learn how we want walls to be painted before it would bake a cake.
The agent must act given reward uncertainty, but maintain its ability to optimize any of the potential rewards in the future when it will know more about the reward. Put simply, it needs to preserve its \emph{option value} while still obtaining as much reward as it can currently~get. 

\begin{figure*}[ht]
\centering
\includegraphics[width=0.97\textwidth]{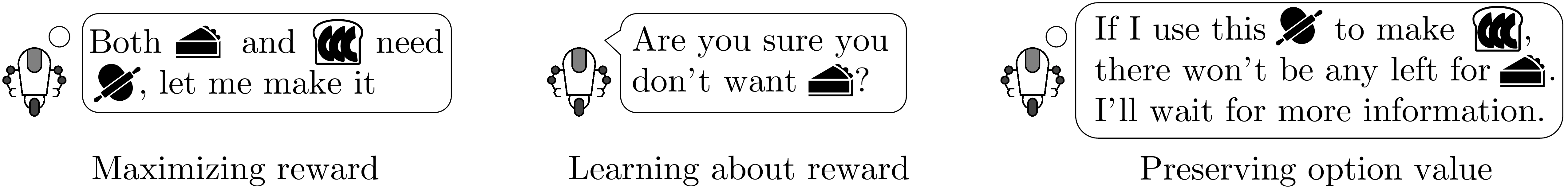}
\caption{In our setting the agent maximizes expected return under reward uncertainty, and can gather more information about the reward. This leads to balancing three considerations: maximizing the reward, learning about the true reward, and preserving option value.}
\label{fig:uses}
\end{figure*}

We would like to evaluate our reward distribution by plugging it into an algorithm that maximizes expected reward given reward uncertainty, but unfortunately current methods~\citep{hadfield2016cooperative, woodward2019learning} are computationally expensive and only work in environments where option value preservation is not a concern. \lawrence{I'm confused. If you train an RNN to maximize reward given uncertainty, it'll preserve option value while it's learning, right? I don't think you need to justify analyzing them separately, tbh.} So, we analyze reward learning and option value preservation separately (Figure~\ref{fig:uses}).

\prg{Active reward learning}
An agent that learns rewards while acting will need to select queries with care. \emph{Active reward learning} aims to pick queries that maximize the information gained about the reward, or minimize the expected regret~\cite{dorsa2017active, mindermann2018active}\footnote{We speculate that future active reward learning methods will also have to ensure that the queries are \emph{relevant} and \emph{timely}: in our household robot example, this would prevent the robot from asking about wall paint color when it is meant to be baking a cake.}. We analyse how the reward distributions resulting from different reward combination methods fit as priors for the active reward learning methods.

\prg{Option-value preservation}~\label{sec:aup}\herke{not sure if the concepts becomes clear enough here. Maybe worth explaining the main concept in a bit more detail?}\lawrence{I think you want a better transition here.} Low impact agents~\citep{turner2019conservative,krakovna2018penalizing,armstrong2017low} take fewer irreversible actions, and so preserve their ability to optimize multiple reward functions. However, work in this area does not consider the possibility of learning more information about the reward function, and so these methods penalize entire \emph{plans} that would have a high impact. In contrast, we would like our agent to pursue impactful plans, but pause when option value would be destroyed, and get more reward information. \herke{This idea of penalizing plans instead of single actions is also mentioned just below Eq. (1), it seems a bit redundant?}

We start with the \emph{attainable utility preservation} (AUP) method introduced by~\citet{turner2019conservative}. AUP formalizes the impact of an action on a reward function as the change in the Q-value as a result of taking that action (relative to a no-op action $\emptyset$). For our purposes, we only care about cases where we are no longer able to optimize the reward function, as opposed to ones where we are \emph{better} able to optimize the reward function. So, we use the truncated difference summary function proposed by~\citet{krakovna2018penalizing} that only counts \emph{decreases} in Q-values and not increases:
\begin{align}
        \text{Impact}(s,a) &= \mathbb{E}_{r}[\max(Q_{r}(s,\emptyset) - Q_{r}(s,a), 0)].
\end{align}
\citet{turner2019conservative} penalize the agent whenever it would cause too much impact, using a hyperparameter $\lambda$ to trade off task reward with impact. They plan using the reward $R_{\text{task}} - \lambda \ \text{Impact}(s,a)$. However, by penalizing impact in the reward function, the agent is penalized for any \emph{plan} that would cause impact, whereas we want our agent to start on potentially impactful plans, but stop once a particular \emph{action} would destroy option value. So, we only penalize impact during action selection:
\begin{align}
    \pi(s) = \max_a (Q(s, a) - \lambda \ \text{Impact}(s,a)). \label{eq:adding_aup_reward}
\end{align}

\herke{is it clear to reviewers what is the practical difference between adding the impact to the Q fc vs the reward function. Also I think the notation for Rtask and Impact wasn't introduced formally?}


\subsection{Acting in a new environment}

Often, we wish to use the reward function in the same environment as the one in which it was specified or inferred. For example, a human expert may have demonstrated how to bake a cake in a particular kitchen. Since we know \emph{at training time} that the agent will be deployed in the same kitchen, the only guarantee we need is that the agent's behavior is correct. If two rewards lead to the same behavior in this kitchen, we don't care which one we end up using.

However, we may instead want to apply a reward function learned in one environment to a different environment. 
In this case, there may be new options in the test environment that we never saw during training. As a result, even if two rewards led to identical behavior in the training environment, they need not do so in the test environment. 
Hence when the goal is to transfer the reward distribution to a new environment, it is crucial to ensure that the distribution contains the true reward, instead of just containing a reward that gives rise to the same behavior as the true one in the training environment.


\section{Desiderata for the reward posterior}\label{chp:desiderata}



We discuss properties of the posterior that inform the choice of a reward combination algorithm for different applications of the resulting reward distribution. We focus on the two applications identified above: option-value preserving planning and active reward learning. 


\subsection{Robust to plausible modes of misspecification}

All else equal, we want the true reward to have high probability in the reward distribution. Below we introduce two plausible ways of misspecifying reward functions linear in features of the state, and formulate two desiderata that give a degree of robustness against these types of misspecification. 

\textbf{Independent per-feature corruption.} One model for misspecification is that the weight for each feature has some chance of being ``corrupted'', and the chance of the feature being corrupted in one input reward is independent from that of the other reward.
If the fraction of misspecified values is small, then if two reward vectors' values are different at a particular index, it is likely that one of the values is correct. \herke{this sentence seems to imply the reverse implication from Desideratum 1, more like "if it is not in indep. points, it should be 0", whereas Des. 1 is "if it is in indep. points, it should be non-0. Very different statements.} Given this, it is desirable for the reward distribution to assign significant probability to every reward function s.t. the value of the weight for each feature is in one of the values of that feature's weight of the input rewards. Formally, we denote the set of such reward parameters 
\begin{align*} \label{eq:indep-features-set}
   \tindep = \{ \theta \in \mathbb R^n \mid \theta_i \in \{\tin{1}_i, \tin{2}_i \} \ \forall i \in \{ 1,...,n\} \}. 
\end{align*}

\herkenew{Probably not actionable for current submission, but based on the intuition above I would expect the opposite desideratum: theta NOT in indep. points should NOT be in the support. Whereas theta in indep points COULD be in support but don't need to be.}
\begin{desideratum}[Support on $\tindep$]\label{des:indep_features}
 All $\tout \in \tindep$ should be in the support of $\posterior$.
\end{desideratum} 



Note that when we do not need to transfer the reward function to a new environment, Desideratum~\ref{des:indep_features} can be made weaker. We only require that the distribution have support on rewards that give rise to the same \emph{behavior} as the rewards in $\tindep$, and not necessarily on $\tindep$ itself.

Des.~\ref{des:indep_features} is helpful for both option value preservation and active reward learning. If it is in fact the case that several features are corrupted independently, having significant probability mass on rewards in $\tindep$ would ensure the agent preserves its ability to optimize them, and would be able to easily learn the true reward with active learning.




\textbf{Misspecified relative importance of different behaviors.} Given a reward vector $\theta$ we define the \textit{tradeoff point} between the i-th and j-th features of the state as the ratio between the values of $\theta$ corresponding to these features: $T_{i,j}^\theta = \frac{\theta_i}{\theta_j}$. The true reward's tradeoff point may be in between the tradeoff points of the input rewards, and so we want our distribution to support all such~tradeoff~points. 


\begin{desideratum}[Support on intermediate feature tradeoffs]\label{des:tradeoffs}
    The reward posterior should have support on rewards that prescribe all intermediate tradeoff points between the states' features.
\end{desideratum}

In practice tradeoff points only affect the behavior to the extent allowed by the environment. 
We only need to maintain \textit{all} intermediate tradeoff points if we want to transfer to a new environment. 

Des.~\ref{des:tradeoffs} is especially important for active reward learning.
Consider a scenario where a personal assistant has to book a flight for its owner Alice, and chooses between options with different costs and durations. \herkenew{Very minor, but wouldn't it be a more sensible example if we made it \$100, \$200 or \$1k, \$2k?} The available flights are (\$1, 9h), (\$2, 5h), (\$6, 2h) and (\$9, 1h). Alice said she wants the shortest flight, giving rise to a reward $\tin{1}=[-1, -9]$. In addition, the assistant observed that previously Alice usually preferred cheaper flights, resulting in the second reward $\tin{2}=[-9,-1]$. Here the assistant is uncertain about Alice's true preference about the tradeoff between cost and time, and so it is sensible to have probability mass on reward functions that would result in choosing any one of the flights, and clarifying this with Alice before making the decision. \dmkr{Highlight that just support on $\tindep$ won't cut it here. Also highlight that we could go all the way to supporting every reward function, but then informativeness won't work}




\subsection{Informative about desirable behavior}
In order for the agent to act, the reward distribution must provide information about useful behavior:

\begin{desideratum}[Informative about desirable behavior] \label{des:informative}
    If $\tin{1}$ and $\tin{2}$ give rise to the same behavior, most of the reward functions in the distribution should give rise to this behavior.
\end{desideratum}
This desideratum can be formalized in different degrees of strength. The strong version states that if the FEs of $\{\tin{1}, \tin{2}\}$ are equal on some feature, FEs of this feature of \textit{any} reward in the support of the posterior equal those of $\{\tin{1}, \tin{2}\}$:
\begin{align}
     &\forall i \in \{1, n \}: \big(
     \fe{\tin{1}}_i =  \fe{\tin{2}}_i  \Rightarrow \nonumber \\   \forall \tout \in &\supp(\posterior):  \fe{\tout}_i = \fe{\tin{1}}_i
     \big).
\end{align}

The medium version considers the FEs as a whole instead of per-feature:
\begin{align}
     &\fe{\tin{1}} = \fe{\tin{2}} \Rightarrow \nonumber \\ \forall \tout \in \supp(&\posterior): \fe{\tout} = \fe{\tin{1}}.
\end{align}

The weak version assumes that the FEs of all $\theta \in \tindep$ are the same, rather than just $\tin{1}$ and $\tin{2}$:
\begin{align}
    \left( \forall \theta \in \tindep: \fe{\theta} = \fe{\tin{1}}= \fe{\tin{2}}
    \right) \Rightarrow \nonumber
    \\  \forall \tout \in \supp(\posterior): \fe{\tout} = \fe{\tin{1}}.
\end{align}


Des.~\ref{des:informative} allows option-value preserving planning to do useful things instead of performing the no-op action all the time: more certainty about the desirable behavior allows it to care less about preserving the expected return attainable by behaviors that are not desirable. 
The strong version of Des.~\ref{des:informative} is especially useful, as it is easy to imagine scenarios where the two input rewards agree on some but not all aspects of the desired behavior. On the other hand, the medium and weak versions of Des.~\ref{des:informative} require the behaviors arising from the input rewards to be exactly the same for all features -- a circumstance we expect to be uncommon in real situations.

Furthermore, by restricting possible behaviors, Des.~\ref{des:informative} simplifies the job of approaches that actively learn the reward function. If the reward distribution used by our household robot always incentivizes using soap while washing dishes (because both input rewards did so), the robot's active learning algorithm does not need to query the human about whether or not to use soap.

Des.~\ref{des:informative} might be problematic if both input rewards agree on an undesirable behavior w.r.t. a particular feature. This presents a natural tradeoff between informativeness and the abilities to preserve option value and actively learn arbitrary reward functions: the more informative we assume the input rewards are about the desired behavior, the less robust we are to their misspecification. 

\subsection{Behavior-space balanced}


Suppose we would like to preserve the option value of pursuing two different behaviors. Option-value preservation is more robust when the distribution over FEs induced by the reward distribution places significant and approximately equal weight on the two behaviors. Intuitively, the ``weight'' AUP puts on preserving expected return attainable by a particular behavior is proportional to the probability of that behavior in the reward distribution. So we would like the FEs of the input rewards to be \textit{balanced} across the reward distribution. Concretely:

\begin{desideratum}[Behavior-space balance]\label{des:behavior-balance}
The FEs of the rewards sampled from the distribution should correspond to behaviors arising from the input reward functions with a similar frequency:
\begin{equation}
    p(\fe{\theta} = \fe{\tin{1}}) \approx p(\fe{\theta} = \fe{\tin{2}})  \mid \theta \sim p(\theta \mid \tin{1}, \tin{2}).
\end{equation}
\end{desideratum}
\herke{It's maybe not so obvious for someone not involved in the project that the above is a constraint on p(theta|...). }





\section{Multitask Inverse Reward Design}

\herke{I found the section up to theorem 1 somewhat hard to follow, but I don't really have concrete suggestions unfortunately. }
Inspired by Inverse Reward Design~\cite{hadfield2017inverse}, we introduce a reward combination method \emph{Multitask Inverse Reward Design} (MIRD). We assume that the two input rewards meet the IRD assumption for two different rewards, namely that the input rewards generate nearly optimal behavior in the training environment according to the corresponding task rewards.
A straightforward way to construct the distribution for the combined reward function is to require the FEs of the rewards in the distribution to be convex combinations of the FEs of the input rewards:
$
    \fe{\tout} \in  \{b \fe{\tin{1}} + (1-b)\fe{\tin{2}} \mid b \in [0,1]\}.
$
\herke{this looked a bit weird to me, as we don't have a single theta* but a distribution. Do we want this to hold for any theta* in the support of the 'posterior'? Or let's say in expectation under the posterior?}


To satisfy this requirement, we propose a distribution that depends on the behavior (set of trajectories) $D = \{\tau^1, ..., \tau^d \}$ resulting from the agents optimizing $\tin{1}$ and $\tin{2}$:
\begin{align}
    p(\tout \mid \tin{1}, \tin{2}) &= \int_D p(\tout \mid D)p(D \mid \tin{1}, \tin{2})  \ dD
\end{align}
Each of the reward functions $\{ \theta^i \mid i=\{1,...,d \} \}$ used to generate the trajectories $\tau^i$ in $D$  is sampled from the Bernoulli mixture over the input reward functions. We introduce a random variable $b\in [0,1]$ that determines the probability of sampling $\theta^i = \tin{1}$ to generate trajectory $\tau^i$:  \herke{make explicit that we are approximating the integrals by sampling!} 
\begin{align}
    p(D \mid \tin{1}, \tin{2}) &= \int_b \Beta(b \mid \beta_1, \beta_2) 
    \prod_{i=1}^n p(\tau^i \mid b) db 
    \\p(\tau^i \mid b) &= b p(\tau^i \mid \tin{1}) + (1-b)p(\tau^i \mid \tin{2}) \nonumber
\end{align}


We sample from $p(\tau^i \mid \theta^i)$ using policies arising from soft value iteration. 
For $p(\tout \mid D)$, we set $\tout$ to the reward learned by Maximum Causal Entropy Inverse RL (MCEIRL)~\citep{ziebart2010modeling}. Denoting the Dirac delta as $\delta()$, we have
$p(\tout \mid D) = \delta(\tout - \argmax_\theta p(\theta \mid D)).$ Having defined all relevant conditional distributions, we can sample from the joint $p(\tout, b, D, \theta^i \mid \tin{1}, \tin{2})$ and the marginal $\posterior$. 

In summary, to generate a sample $\tout$ from $\posterior$, we (1) simulate $D$, a set of trajectories in which $b\times100$\% trajectories arise from $\tin{1}$ and the rest from $\tin{2}$, and (2) do MCEIRL with $D$.

\begin{theorem}\label{thm:mird-behavior}
Given the rationality parameter $k$ used in MCEIRL, soft value iteration with rationality $k$ applied to the rewards $\tout$ in the support of the MIRD posterior results in FEs that are convex combinations of FEs of the input rewards $\tin{1}$ and $\tin{2}$:
\begin{align}
    &\forall \tout \in \supp(\posterior):\nonumber\\  
    \fe{\tout} &= \alpha \fe{\tin{1}} + (1- \alpha) \fe{\tin{2}}, \ \alpha \in [0,1].
\end{align}
\end{theorem}
\textit{Proof.} See appendix 
A for full proof. Key to the proof is the fact that MCEIRL finds such $\tout$ that its FEs match those of the soft value iteration expert who generated the trajectories~\cite{ziebart2010modeling}. In our case, the FEs of $\tout$ match $D$, which contains trajectories arising from $\tin{1}$ and $\tin{2}$ in various proportions. 

While Theorem~\ref{thm:mird-behavior} considers soft value iteration, note that this can be made arbitrarily close to regular value iteration by making the rationality parameter $k$ arbitrarily large.

\begin{corollary}
The average FEs of the reward functions in the posterior are:
\begin{align}\label{eq:mird-average-fes}
    \expect{p(\theta \mid \tin{1}, \tin{2})}{\fe{\theta}} 
    &=
    \expect{b}{b}  \fe{\tin{1}} + (1 - \expect{b}{b}) \fe{\tin{2}}.
\end{align}
\end{corollary}



\begin{corollary}
Eq.~\ref{eq:mird-average-fes} implies that if the FEs of the $\tin{1}$ and $\tin{2}$ w.r.t. a given feature are equal, the FEs of all rewards in the distribution for that feature equal the FEs of $\tin{1}$ and $\tin{2}$ for that feature. Thus MIRD satisfies the strong version of Des.~\ref{des:informative} (informative about desirable behavior).
\end{corollary}

\begin{corollary}
When $\beta_1 = \beta_2$ the probability of sampling $\theta^i = \tin{1}$ equals the probability of sampling $\theta^i = \tin{2}$, and hence the probabilities of observing behaviors $\mathcal{F}^{\pi_{\tin{1}}}$ and $\mathcal{F}^{\pi_{\tin{2}}}$ when optimizing a reward function from the posterior are equal. Hence Des.~\ref{des:behavior-balance} (behavior balance) is satisfied.
\end{corollary}



Theorem~\ref{thm:mird-behavior} allows us to bound the expected return ${\theta^{\true}}^\top \fe{\theta}$, $\theta\in \supp(\posterior)$ of the true reward $\theta^{\true}$ in terms of expected true returns of $\fe{\tin{1}}$ and $\fe{\tin{2}}$:

\begin{theorem} \label{thm:mird-true-return}
The minimum expected true return achieved by optimizing any of the rewards in the MIRD posterior with soft value iteration with rationality $k$  equals the minimum true return from optimizing one of the input rewards:
\begin{equation}
    \min_{\theta \in \supp(\posterior)} 
    {\theta^{\true}}^\top \fe{\theta} = \min_{\theta \in \{ \tin{1}, \tin{2}\} } {\theta^{\true}}^\top \fe{\theta}.
\end{equation}
\end{theorem}
\textit{Proof.} See Appendix B.

Theorem~\ref{thm:mird-true-return} is the best possible regret bound for following rewards in a distribution resulting from reward combination: \herke{add the reasoning for this claim? Perhaps not obvious why there couldn't be reward combination strategy that does better.}  the worst the agent can do when following a reward from the distribution is no worse than when just randomly choosing to follow one of the input reward functions. \herke{I think you are right about being the best possible bound, but to me it's not obvious, so might need to support this more clearly for reviewers.}


Generally FEs arising from $\theta \in \tindep\backslash \{\tin{1}, \tin{2} \}$ are not convex combinations of $ \mathcal{F}^{\pi_{\tin{1}}} $ and $\mathcal{F}^{\pi_{\tin{2}}}$, so MIRD does not satisfy any version of Des.~\ref{des:indep_features} (support on $\tindep$). 
When $\tin{1}$ and $\tin{2}$ conflict on more than one feature, Des.~\ref{des:tradeoffs} (support on intermediate feature tradeoffs) is not satisfied either. 

\prg{Independent features formulation}\label{sec:mird-if}
Since both Des.~\ref{des:indep_features} and Des.~\ref{des:tradeoffs} aim to ensure robustness to reward misspecification, one of our primary goals, we would like a variant of MIRD that satisfies them. An algorithm that supports rewards that give rise to all potential behaviors arising from reward functions in  $\tindep$ is a simple modification of MIRD. The only change required in the original formulation of MIRD is sampling the reward vectors $\{ \theta^i \}$ giving rise to the trajectories in $D$ from $\tindep$ instead of just $\{ \tin{1}, \tin{2} \}$. Now $b$ is a vector of dimensionality $|\tindep|$, sampled from $\Dirichlet(b \mid \beta_1,...,\beta_{|\tindep|} )$. Each $\theta^i$ is sampled from a Multinoulli distribution over $\tindep$, parameterized by $b$.
We refer to this version of MIRD as MIRD independent features, or MIRD-IF. 



\begin{table*}[ht]
\centering
\caption{Performance of reward combination algorithms on the desiderata outlined in Section~\ref{chp:desiderata}.}
\label{tab:methods-vs-des}
\footnotesize
\begin{tabular}{l|cccccc}
\toprule
 &  \multicolumn{3}{l}{\begin{tabular}[c]{@{}l@{}}Informative about desirable \\  behavior\end{tabular}} & \begin{tabular}[c]{@{}l@{}}Support on \\ $\tindep$\end{tabular} & \begin{tabular}[c]{@{}l@{}}Support on intermediate \\ feature tradeoffs\end{tabular}  & \begin{tabular}[c]{@{}l@{}}Behavior\\ balance\end{tabular} \\
 &   (strong) & (medium) & (weak) & & &  \\ \hline
Additive & \nocross & \yescheck & \yescheck & \nocross & \nocross & \nocross \\ \hline
Gaussian & \nocross & \nocross & \nocross & \yescheck & \yescheck & \nocross \\ \hline
CC-in & \nocross & \yescheck & \yescheck & \nocross & \nocross & \kindof \\ \hline
CC-$\tindep$ & \nocross & \nocross & \yescheck & \yescheck & \yescheck & \kindof \\ \hline
Uniform Points & \nocross & \nocross & \yescheck & \yescheck & \nocross & \yescheck \\ \hline
MIRD & \yescheck & \yescheck & \yescheck & \nocross & \nocross & \yescheck \\ \hline
MIRD-IF & \nocross & \nocross & \yescheck & \yescheck & \yescheck & \yescheck \\ \bottomrule
\end{tabular}
\end{table*}

\section{Analysis} \label{sec:experiments}

We first introduce several baselines each of which does not meet some of the outlined desiderata, and then analyze all reward combination methods in the context of our modified version of AUP.

\subsection{Reward-space baselines}

\textbf{Additive.}\label{sec:additive}
The additive reward combination method is commonly used in previous work utilizing reward combination~\cite{kolter2009near, desai2018negotiable, rlsp}. The reward posterior is simply $p(\tout \mid \tin{1}, \tin{2}) = \delta(\tout - (\tin{1} + \alpha \tin{2}))$,
where $\alpha$ is the hyperparameter used to balance the two input reward functions.

\textbf{Gaussian.} Here, we use the standard Bayesian framework, and assume that $\tin{1}$ and $\tin{2}$ are generated from $\tout$ with Gaussian noise:
\begin{align*}
    p(\tout \mid \tin{1}, \tin{2}) &\propto p(\tout)p(\tin{1} \mid \tout)p(\tin{2} \mid \tout)
    \\&=p(\tout)p_{\mathcal{N}}(\tin{1} \mid \tout, \sigma_1^2\bm I) p_{\mathcal{N}}(\tin{2} \mid \tout, \sigma_2^2\bm I),
\end{align*}
where $\bm I$ is the identity matrix, $p_{\mathcal{N}}$ is the multivariate Gaussian probability density function, and $\{\sigma_1,\sigma_2\}$ are hyperparameters controlling the standard deviations. One can express higher trust in a given input reward by lowering its $\sigma$. 
For our experiments, we use an uninformative Gaussian prior (that is, $\sigma \rightarrow \infty$) and set $\sigma_1 = \sigma_2 = 1$. The posterior is then $p(\tout \mid \tin{1}, \tin{2}) = p_{\mathcal{N}}(\tout \mid \frac{\tin{1} + \tin{2}}{2}, \frac{1}{2}\bm I)$.


\textbf{Convex combinations of input reward vectors (CC-in).}\label{sec:convex-comb-reward}
Consider a scenario where the input reward functions are correctly specified for two different tasks. 
A natural reward-space analogue to MIRD would be to consider a distribution over all convex combinations of the input reward vectors: 
\begin{align}
    \posterior =  \int_b  
    p(b)\delta(\tout - b \tin{1}-(1-b)\tin{2})db.
\end{align}
As before, the weight $b$ is sampled from $\Beta(b \mid \beta_1, \beta_2)$.

\begin{theorem}\label{thm:convex-combs}
If $\fe{\tin{1}} = \fe{\tin{2}}$, the FEs of any reward in the support of the CC-in posterior
equal $\fe{\tin{1}}$. Hence CC-in satisfies the medium version of Des.~\ref{des:informative} (informative about desirable behavior).

Proof. \normalfont{See Appendix C.
}
\end{theorem}

We provide an example showing that CC-in does not satisfy the strong version of Des~\ref{des:informative} in Appendix E.
Furthermore, CC-in does not satisfy Des.~\ref{des:indep_features} (support on $\tindep$): consider combining rewards $[0,1,1]$ and $[1,0,1]$. Here $\theta=[0,0,1] \in \tindep$, but is not a convex combination of the input rewards. 

\textbf{Convex combinations of $\theta \in \tindep$ (CC-$\tindep$).} Instead of only considering convex combinations of the input rewards, we consider convex combinations of $\theta \in \tindep$:
\begin{align}
    \posterior &= \int_b p(b) \delta \bigg( \tout - \sum_{i=0}^{|\tindep|} b_i \tindep_i \bigg)db.
    \label{eq:cc-tindep}
\end{align}
The vector $b \in \mathbb{R}^{|\tindep|}$ of weights for each convex combination is sampled from the Dirichlet distribution: $p(b) = \Dirichlet(b \mid \beta_1,...,\beta_{|\tindep|} )$.


\begin{theorem}\label{thm:convex-combs-indep}
If all FEs of rewards in $\tindep$ are equal, the FEs of any reward in the CC-$\tindep$  posterior equal the FEs of rewards in $\tindep$. Hence CC-$\tindep$ satisfies Des.~\ref{des:informative} (weak).

\textit{Proof.} \normalfont{See Appendix D.
}
\end{theorem}



\textbf{Uniform points.} Note that as $\beta_1 = ... = \beta_{|\tindep|} \rightarrow 0$, Eq.~\ref{eq:cc-tindep} essentially becomes $p(\tout \mid \tin{1}, \tin{2}) = \text{Uniform} (\tindep)$. It is helpful to analyze the properties of this \emph{uniform points} distribution separately.

Table~\ref{tab:methods-vs-des} summarizes the properties of MIRD, MIRD-IF, and the reward-space baselines. \herke{It is maybe not obvious where all the marks come from. I think the ones from MIRD are pretty explicit but not for all the other ones like MIRD-IF.} 



\subsection{Analysis on a toy environment}
\herke{given we have more space I would pull more of the experimental results forwards...} To demonstrate the use of the desiderata and each method's performance on them, we developed a toy environment for testing reward combination methods. Due to space limitations, we defer the full qualitative and quantitative explanations to Appendix F 
and report the broad results here.

\begin{wrapfigure}[26]{h}{0.28\textwidth}
\centering
\includegraphics[width=0.24\textwidth]{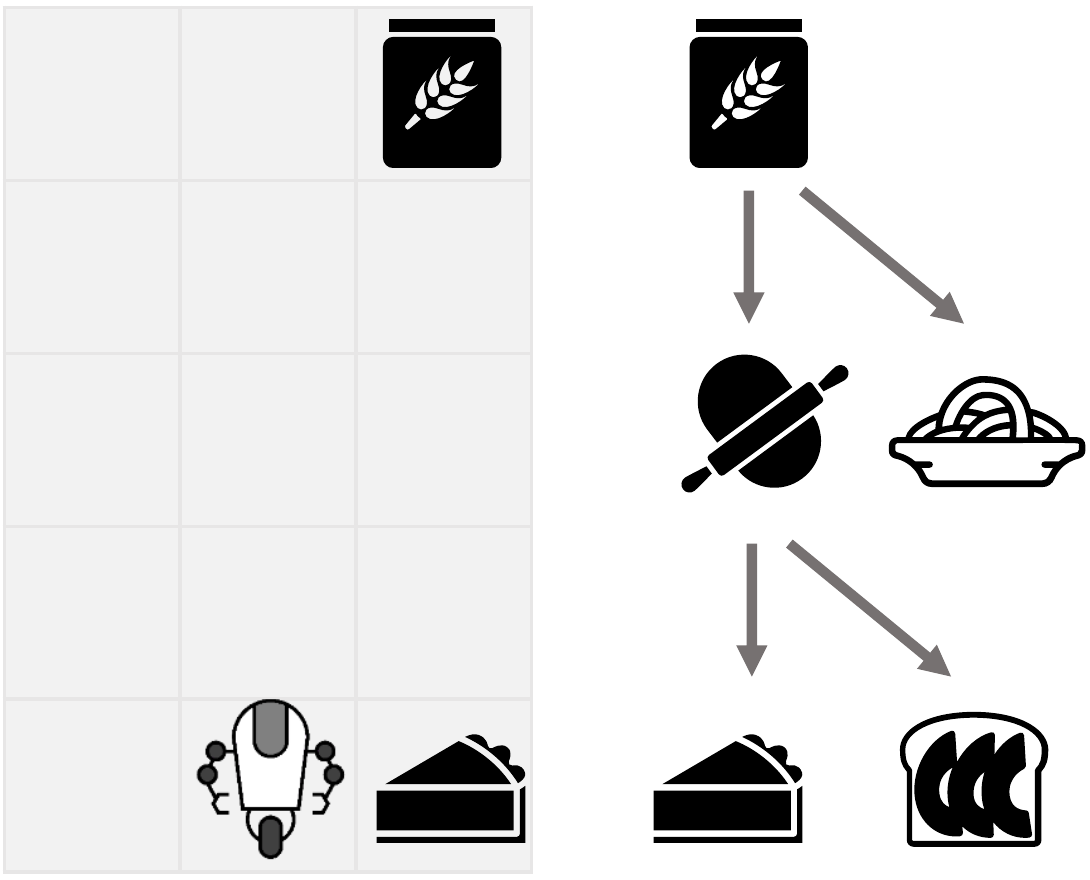}
\caption{The Cooking environment. Left: the gridworld layout and the current state. The states' features indicate the number of jars of flour, pieces of dough, cakes, avocado toasts, and servings of pasta.
Right: irreversible actions for cooking. At each state the agent can move, cook, or do nothing. Cooking is possible when the agent is to the left of flour or dough. Making cake or avocado toast involves first making dough from the flour, and then making cake or toast; making pasta uses flour directly. 
}
\label{fig:envs}
\end{wrapfigure}

\textbf{Setup.} We use the setting of AUP planning to demonstrate the importance of support on $\tindep$, informativeness about desired the behavior, and behavior balance desiderata for option value preservation. We do not evaluate Des~\ref{des:tradeoffs} (support on intermediate feature tradeoffs), as it is not helpful for preserving option value: it is primarily useful for active learning, which we do not test. We obtain the reward posterior by combining the input rewards using each of our seven reward combination methods, and evaluate the behavior of AUP planning w.r.t. each posterior over five seeds. Our implementation of AUP uses state-action value functions computed with traditional value iteration with discount $\gamma=0.98$; the impact penalty weight $\lambda$ is set to 1 unless specified otherwise. We use $\beta=0.5$ for the Beta and the Dirichlet distributions used in CC-in, CC-$\tindep$, MIRD, and MIRD-IF. The input rewards are either both specified by hand, or $\tin{1}$ is learned with RLSP while $\tin{2}$ is hand-specified. 
We use the \textit{Cooking} environment described in Figure~\ref{fig:envs}. 

\textbf{Support on $\tindep$.} We specify both input rewards to strongly incentivise either cake or toast by setting $\tin{1}_{\text{cake}} = 3, \tin{1}_{\text{toast}} = 0$ and $\tin{2}_{\text{cake}} = 0, \ \tin{2}_{\text{toast}} = 3$, 
and to mildly incentivise pasta: $\tin{1}_{\text{pasta}} = \tin{2}_{\text{pasta}} = 1$. The desired behavior is to preserve the flour, as any one of the foods might be desirable. This requires the distribution to place weight on rewards such as  $\theta_{\text{cake}} = \theta_{\text{toast}}=0$; $\theta_{\text{pasta}}=1$ which would make pasta. As expected, methods that have support on $\tindep$ behave as desired, and other methods do not.
\herkenew{here (and with the other paragraph after this) as a reviewer I'd like to see the exact, `quantitative' results. If other methods do not, what do they do instead? Does it make dough, or do nothing, or make toast? Does it happen 5/5 random seeds or only sometimes? Maybe it takes to much space to discuss here but then you could move it to appendix?}

\textbf{Informativeness about desirable behavior (strong).} In the setting above the two input rewards lead the agent to make either cake or toast, but not pasta. So, a reward distribution meeting the strong version of Des.~\ref{des:informative} would only make cake or toast, and so the agent should make dough (which is useful for both). \herke{would only make... sounds like a hypothetical. Maybe "did only make" or similar?} Note that despite the same setup, this is exactly the opposite behavior than desired previously, showing the tradeoff between informativeness and support on $\tindep$. As such, our experimental results are reversed.
\herke{maybe note the inherent conflict between independent points and informativeness earlier in the text? So it doesn't seem weird here...}


\textbf{Informativeness about desirable behavior (weak).} We hand-specify one of the reward functions to incentivise the agent to make cakes: $\tin{2}_{\text{cake}}=1,\ \tin{2}_{\text{toast}}=\tin{2}_{\text{pasta}}=0$. $\tin{1}$ is inferred by RLSP with a uniform prior over $s_{-T}$. Since the current state contains a cake, RLSP infers a positive reward for cakes, $\tin{1}_{\text{cake}}\approx0.1$, and near-zero rewards for the other features. 
Hence we have two rewards that both incentivise the agent to make another cake. The desired behavior is to make the cake, and not worry about preserving the ability to make pasta or toast as neither input reward cares about it. All but the Gaussian method satisfy the weak informativeness desideratum, and so they all succeed. The Gaussian method fails since it thinks $\tout_{\text{cake}}$ could be negative.


\textbf{Behavior balance.} 
$\tin{2}$ is specified to only reward toast, $\tin{2}_{\text{toast}}=1$, $\tin{2}_{\text{cake}}=\tin{2}_{\text{pasta}}=0$, and $\tin{1}$ is specified only reward cake: $\tin{1}_{\text{cake}}>0$, $\tin{1}_{\text{toast}}=\tin{1}_{\text{pasta}}=0$. So we have two disagreeing rewards: one incentivises cake, while the other incentivises toast. The desired behavior from AUP here is to make dough as it is desirable for both the cake and the toast, but then to stop as the next step is unclear. Ensuring that the reward for toast does not overwhelm the reward for cake requires behavior balance. We vary $\tin{1}_{\text{cake}}$ between 0.01 and 0.7 to obtain many degrees of imbalance between toast and cake, and vary the AUP weight $\lambda$ between 0.01 and 5 to see the effects of behavior imbalance on AUP. We plot our results in appendix F. 
MIRD and MIRD-IF have the lowest frequency of the toast reward overwhelming the cake reward, which indicates that MIRD and MIRD-IF are most behavior balanced. However, occasionally these methods do nothing instead of making dough.



\section{Related Work}\label{sec:related}




\textbf{Utility aggregation.} Combining the specified reward functions is in many ways similar to combining utility functions of different agents. Harsanyi’s aggregation theorem~\cite{harsanyi1955cardinal} suggests that maximizing a fixed linear combination of utility functions results in Pareto-optimal outcomes for correctly specified utility functions. \rs{Isn't it that all Pareto-optimal outcomes correspond to some fixed linear combination of utility functions?} This corresponds to our Additive baseline. 

\textbf{Multi-task and meta  inverse RL.}  Multi-task IRL assumes that the demonstrations originate from multiple experts performing different tasks, and seeks to recover reward functions that explain the preferences of each of the experts \citep{babes2011apprenticeship,choi2012nonparametric}. 
The IRL part of MIRD and MIRD-IF is conceptually similar to these approaches: the trajectories in each sampled demonstration set $D$ can be seen as arising from an expert executing different tasks. However, MIRD seeks to explain each $D$ as if the trajectories arose from a \textit{single} reward function. This resembles meta IRL: \citet{li2017meta} explain the preferences of an expert who demonstrated several different tasks by learning a reward shared between the tasks, while \citet{xu2018learning} and \citet{gleave2018multi} learn an initialization for the reward function parameters that is helpful for learning the rewards of new tasks. MIRD is different from these methods in that it (1) uses reward functions instead of expert demonstrations as inputs, (2) recovers the full distribution over the underlying reward function, and (3)~explicitly accounts for misspecification.






\section{Limitations and future work}

\textbf{Summary.} We analyze the problem of combining two potentially misspecified reward functions into a distribution over rewards. We identify active reward learning and option value preservation as two key applications for such a distribution, and determine four properties that the distribution would ideally satisfy for these applications. 
We suggest some simple reward-space methods as well as a behavior-space method, Multitask Inverse Reward Design (MIRD), which is grounded in the behavior arising from the input rewards. MIRD works well for most applications, and MIRD-IF can be used for greater robustness and increased likelihood that the true reward is in the distribution. If the only application is to preserve option value, then the uniform points method also works well.


\textbf{Active reward learning and refined desiderata.} The primary avenue for future work is to evaluate the reward combination methods in the context of active reward learning, and formalize Des.~\ref{des:tradeoffs} (support on intermediate tradeoffs) in a way most helpful for active learning. Furthermore, it would be helpful to introduce a soft version of Des.~\ref{des:informative} (informativeness), as the current formulation only permits the reward distribution to have support on a subspace of $\mathbb{R}^{n}$ instead of supporting the entire $\mathbb{R}^{n}$, as preferable for active learning.

\textbf{Realistic environments.} Another avenue for future work is scaling up MIRD to realistic environments in which the dynamics are not known, the state space is not enumerable, and the reward function may be nonlinear. While it would be straightforward to use an existing deep IRL algorithm \cite{fu2017learning, finn2016guided} in place of MCEIRL, running a maximum likelihood IRL algorithm to generate a single sample from the reward distribution could be prohibitively expensive. Instead, it might be helpful to look into using Bayesian Neural Networks~\cite{mackay1992practical} to model the reward posterior.

\textbf{Other sources of preference information.} Future work could use MIRD to not only combine reward functions, but to combine the preference information encoded in policies or even trajectories: for example, it would be straightforward to use an input policy (obtained e.g. with imitation learning) in place of one of the input reward functions to generate the trajectories in $D$.

\lawrence{I think there's a lot more to say here. For example, there's categorizing how common misspec issues affect the reward you get as well as no free lunch theorems (theory), more experiments in less toy environments (gridworlds, mujoco), and also experiments with human models/real humans.}

\bibliographystyle{apalike}
\bibliography{references}

\appendix
{
\section{Proof of Theorem~1}\label{proof:mird-fes}

$D$ is a set of trajectories each of which arises either from optimizing $\tin{1}$ or optimizing $\tin{2}$ with soft value iteration with rationality $k$. The fraction of trajectories in $D$ that result from $\tin{1}$ is $b$, and the fraction of trajectories arising from $\tin{2}$ is $1-b$. Hence the feature expectations (FEs) of $D$ are
\begin{align}\label{eq:mird-proof-convex}
    \fe{D} = b \fe{\tin{1}} + (1-b) \fe{\tin{2}}.
\end{align}

$p(\tout \mid D) = \delta(\tout - \argmax_\theta p(\theta \mid D))$ entails that each sample $\tout$ is the MCEIRL reward $\tout= \argmax_\theta p_{\text{MCEIRL}}(\theta \mid D)$. MCEIRL with rationality $k$ finds a reward vector $\tout$ such that its FEs match those of the soft value iteration expert (with rationality $k$) who generated the trajectories~\cite{ziebart2010modeling}. Hence each sampled $\tout$ gives rise to FEs $\fe{\tout}$ matching $D$. This and Eq.~\ref{eq:mird-proof-convex} together imply that the FEs arising from optimizing any $\tout \in \supp(\posterior)$ with soft value iteration are a convex combination of the FEs of the input rewards:
\begin{align}
    \forall \tout \in \supp(\posterior): \ \fe{\tout} = b \fe{\tin{1}} + (1- b) \fe{\tin{2}}, \ b \in [0,1]. \ \blacksquare
\end{align}

\section{Proof of Theorem~2}\label{proof:mird-true-return}
By Theorem~1, the FEs arising from any $\theta$ in the posterior are a convex combination of the FEs of the input rewards. Hence finding $\theta$ in the support of $p(\tout \mid \tin{1}, \tin{2})$ that minimizes ${\theta^{\true}}^\top \fe{\theta}$ is equivalent to finding $\alpha \in [0,1]$ that minimizes
\begin{align*}
    {\theta^{\true}}^\top (\alpha \fe{\tin{1}} + (1- \alpha) \fe{\tin{2}}) 
    = \alpha {\theta^{\true}}^\top \fe{\tin{1}} + (1- \alpha) {\theta^{\true}}^\top \fe{\tin{2}}.
\end{align*}
Since the true returns when following policies arising from the input reward functions ${\theta^{\true}}^\top  \fe{\tin{1}}$ and ${\theta^{\true}}^\top \fe{\tin{2}}$ are scalars, the term above is minimized with $\alpha=1$ when ${\theta^{\true}}^\top  \fe{\tin{1}}<{\theta^{\true}}^\top \fe{\tin{2}}$, and $\alpha=0$ otherwise. $\blacksquare$

\section{Proof of Theorem~3} \label{proof:convex-combs}
Since the FEs of the input rewards are equal, the return of any mixture reward $b \tin{1}-(1-b)\tin{2}$ for those FEs is $\fe{\tin{1}}(b \tin{1}-(1-b)\tin{2})$. Suppose there exist FEs $\mathcal F'$ that result in a higher expected return for the mixture reward:
\begin{equation}
    \mathcal{F}'(b \tin{1}-(1-b)\tin{2}) > \fe{\tin{1}}(b \tin{1}-(1-b)\tin{2})
\end{equation}
For this to be true $\mathcal{F}'$ must result in a higher return than $\fe{\tin{1}}$ for either $\tin{1}$ or $\tin{2}$. This is a contradiction, since $\fe{\tin{1}}$ already results in the highest return for both input rewards. Hence $\fe{\tin{1}}$ corresponds to the optimal behavior for any mixture reward $b \tin{1}-(1-b)\tin{2} \ \forall b \in [0,1]$.   $\blacksquare$ 

\section{Proof of Theorem~4} \label{proof:convex-combs-indep}

Analogous to the proof of Theorem~3 (Appendix~\ref{proof:convex-combs}).
}

\section{Additive and CC-in do not satisfy the strong version of Des.~3 (informativeness)}\label{appendix-additive-ccin}

\begin{figure}[ht]
\centering
\tikzset{every picture/.style={line width=0.75pt}} 

\begin{tikzpicture}[x=0.75pt,y=0.75pt,yscale=-1,xscale=1]

\draw   (58.4,68.83) .. controls (58.4,60.1) and (65.47,53.03) .. (74.2,53.03) .. controls (82.93,53.03) and (90,60.1) .. (90,68.83) .. controls (90,77.55) and (82.93,84.63) .. (74.2,84.63) .. controls (65.47,84.63) and (58.4,77.55) .. (58.4,68.83) -- cycle ;

\draw   (120,28.83) .. controls (120,20.1) and (127.07,13.03) .. (135.8,13.03) .. controls (144.53,13.03) and (151.6,20.1) .. (151.6,28.83) .. controls (151.6,37.55) and (144.53,44.63) .. (135.8,44.63) .. controls (127.07,44.63) and (120,37.55) .. (120,28.83) -- cycle ;

\draw    (140,13.03) .. controls (150.78,-11.25) and (176.61,2.35) .. (151.6,21.83) ;
\draw [shift={(150,23.03)}, rotate = 323.99] [color={rgb, 255:red, 0; green, 0; blue, 0 }  ][line width=0.75]    (10.93,-3.29) .. controls (6.95,-1.4) and (3.31,-0.3) .. (0,0) .. controls (3.31,0.3) and (6.95,1.4) .. (10.93,3.29)   ;

\draw   (120,68.83) .. controls (120,60.1) and (127.07,53.03) .. (135.8,53.03) .. controls (144.53,53.03) and (151.6,60.1) .. (151.6,68.83) .. controls (151.6,77.55) and (144.53,84.63) .. (135.8,84.63) .. controls (127.07,84.63) and (120,77.55) .. (120,68.83) -- cycle ;

\draw    (140,53.03) .. controls (150.78,28.75) and (176.61,42.35) .. (151.6,61.83) ;
\draw [shift={(150,63.03)}, rotate = 323.99] [color={rgb, 255:red, 0; green, 0; blue, 0 }  ][line width=0.75]    (10.93,-3.29) .. controls (6.95,-1.4) and (3.31,-0.3) .. (0,0) .. controls (3.31,0.3) and (6.95,1.4) .. (10.93,3.29)   ;

\draw   (120,108.83) .. controls (120,100.1) and (127.07,93.03) .. (135.8,93.03) .. controls (144.53,93.03) and (151.6,100.1) .. (151.6,108.83) .. controls (151.6,117.55) and (144.53,124.63) .. (135.8,124.63) .. controls (127.07,124.63) and (120,117.55) .. (120,108.83) -- cycle ;

\draw    (140,93.03) .. controls (150.78,68.75) and (176.61,82.35) .. (151.6,101.83) ;
\draw [shift={(150,103.03)}, rotate = 323.99] [color={rgb, 255:red, 0; green, 0; blue, 0 }  ][line width=0.75]    (10.93,-3.29) .. controls (6.95,-1.4) and (3.31,-0.3) .. (0,0) .. controls (3.31,0.3) and (6.95,1.4) .. (10.93,3.29)   ;

\draw    (90,68.83) -- (118.8,107.23) ;
\draw [shift={(120,108.83)}, rotate = 233.13] [color={rgb, 255:red, 0; green, 0; blue, 0 }  ][line width=0.75]    (10.93,-3.29) .. controls (6.95,-1.4) and (3.31,-0.3) .. (0,0) .. controls (3.31,0.3) and (6.95,1.4) .. (10.93,3.29)   ;

\draw    (90,68.83) -- (118,68.83) ;
\draw [shift={(120,68.83)}, rotate = 540] [color={rgb, 255:red, 0; green, 0; blue, 0 }  ][line width=0.75]    (10.93,-3.29) .. controls (6.95,-1.4) and (3.31,-0.3) .. (0,0) .. controls (3.31,0.3) and (6.95,1.4) .. (10.93,3.29)   ;

\draw    (90,68.83) -- (118.8,30.43) ;
\draw [shift={(120,28.83)}, rotate = 486.87] [color={rgb, 255:red, 0; green, 0; blue, 0 }  ][line width=0.75]    (10.93,-3.29) .. controls (6.95,-1.4) and (3.31,-0.3) .. (0,0) .. controls (3.31,0.3) and (6.95,1.4) .. (10.93,3.29)   ;

\draw (135.8,28.83) node   {$s_{2}$};
\draw (135.8,68.83) node   {$s_{3}$};
\draw (74.2,68.83) node   {$s_{1}$};
\draw (135.8,108.83) node   {$s_{4}$};
\draw (229.5,30.53) node [scale=0.9]  {$f( s_{2}) \ =[ 1,0,1] \ $};
\draw (229.5,70.53) node [scale=0.9]  {$f( s_{3}) \ =[ 0,1,1] \ $};
\draw (241,110.53) node [scale=0.9]  {$f( s_{4}) \ =[ 0.9,0.9,0] \ $};
\draw (53.5,30.53) node [scale=0.9]  {$f( s_{1}) \ =[ 0,0,0] \ $};

\end{tikzpicture}
\caption{Toy reward combination problem. The aim is to combine reward vectors $\tin{1} = [1,0,0]$ and $\tin{2}=[0,1,0]$. The MDP is as follows. The starting state is $s_1$. From $s_1$ the agent must irreversibly and deterministically transition to $s_2$, $s_3$, or $s_4$, and stay there indefinitely. The features of each state $s$ are given by $f(s)$. } \label{fig:mdp-4-state-2}
\end{figure}
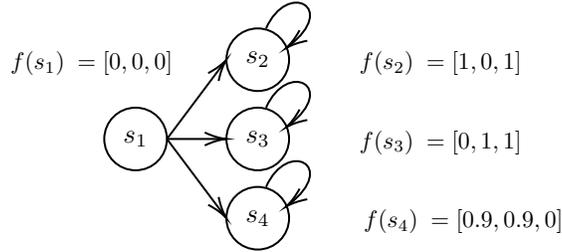

\prg{Additive} Consider the reward combination scenario in Figure~\ref{fig:mdp-4-state-2}. Maximizing $\tin{1}$ or $\tin{2}$ an agent would correspondingly end up in $s_2$ or $s_3$, both of which have the third feature equal 1. However, maximizing $\tout = \tin{1}+\tin{2}$ would lead the agent to $s_4$, where the value of the third feature is 0.

\prg{CC-in} Again consider the reward combination task in Figure~\ref{fig:mdp-4-state-2}. Maximizing $\tin{1}$ and $\tin{2}$ separately would correspondingly lead to choosing $s_2$ and $s_3$, both states having the third feature equal 1. However, maximizing $\tout = \frac{\tin{1}+\tin{2}}{2} = [0.5, 0.5, 0]$, a convex combination of the input rewards, would lead to the agent choosing $s_4$ where the value of the third feature equals 0. 

As the Additive and CC-in reward posteriors have support on rewards giving rise to behaviors \textit{different} from the behaviors of the input rewards w.r.t. the third feature of the toy MDP, neither method meets the strong version of Des.~3 (informative about desirable behavior).

\section{Analysis details} \label{appendix:analysis}

In this appendix we provide more details and explanations regarding the behavior of the various reward combination methods in the Cooking environment. Every qualitative result reported below is validated over five seeds.

\subsection{Support on $\tindep$} We hand-specify both input reward functions to strongly incentivise either cake or toast by setting $\tin{1}_{\text{cake}} = \tin{2}_{\text{toast}} = 3$; $\tin{1}_{\text{toast}} = \tin{2}_{\text{cake}} = 0$, and to mildly incentivise pasta: $\tin{1}_{\text{pasta}} = \tin{2}_{\text{pasta}} = 1$. 
The desired behavior from AUP here is to preserve the flour, as any one of the foods requiring it might be desirable. 


The Additive method returns a single reward function that incentivises the agent to make both cake and toast. AUP applied to one reward function simply optimizes it without needing to preserve option value, so here it makes either cake or toast (and does not preserve the flour).

The CC-in method does not preserve the flour, choosing to make dough instead. Because of the high values of $\tin{1}_{\text{cake}}$ and $\tin{2}_{\text{toast}}$, optimizing any convex combination of $\{\tin{1}, \tin{2}\}$ leads the agent to either make cake or toast, and so AUP infers that it does not need to preserve the ability to make pasta. MIRD chooses to make dough for a similar reason: since all trajectories in $D$ arise from optimizing $\tin{1}$ or $\tin{2}$, each trajectory demonstrates the agent making either cake or toast, and no trajectory demonstrates it making pasta. So no reward inferred from $D$ incentivises the agent to make pasta, and hence AUP does not preserve the ability to make pasta.

All methods whose reward samples sometimes give rise to the behaviors arising from $\theta \in \tindep$ preserve the flour. This is because $\tindep$ contains $\theta$ s.t. optimizing it leads the agent to make pasta: $\theta_{\text{cake}} = \theta_{\text{toast}}=0$; $\theta_{\text{pasta}}=1$. As MIRD-IF, Gaussian, uniform points, and CC-$\tindep$ methods all have substantial probability mass on rewards that incentivise making pasta, AUP applied to the reward samples from these methods preserves the ability to make pasta.

\subsection{Informativeness about desirable behavior (strong).} In the setting above the two input rewards lead the agent to make either cake or toast, but not pasta. Here a reward posterior meeting the strong version of Des.~3 would only give rise to reward functions that incentivise the agent to make either cake or toast. Note that despite the same setup, this is exactly the opposite behavior than desired previously, showing the tradeoff between informativeness and support on $\tindep$. As such, the results are reversed.

We see that MIRD-IF, Gaussian, uniform points, and CC-$\tindep$ methods do not meet Des.~3 (strong), since they all have substantial probability mass on rewards that incentivise making pasta. Because of this AUP preserves the ability to make pasta, and does not lead the agent to make dough. 
On the other hand, MIRD, Additive, and CC-in methods never give rise to rewards incentivising the agent to make pasta, and AUP applied to these reward distributions results in making dough. Although Additive and CC-in do not support reward functions that lead the agent to make pasta in this particular case, they do not meet the strong version of Des.~3 (shown in Appendix~\ref{appendix-additive-ccin}).


\subsection{Informativeness about desirable behavior (medium \& weak)} We hand-specify one of the reward functions to incentivise the agent to make cakes: $\tin{2}_{\text{cake}}=1,\ \tin{2}_{\text{toast}}=\tin{2}_{\text{pasta}}=0$. $\tin{1}$ is inferred by RLSP with a uniform prior over $s_{-T}$. 
\dmkr{motivate RLSP? Tho probably fine as is since there's no vase and stuff; or say something in setup?}
The current state contains a cake which was likely made by a human, so RLSP infers a positive reward for cakes, $\tin{1}_{\text{cake}}\approx0.1$, and near-zero rewards for the other features. 
So we have two agreeing rewards that both incentivise the agent to make another cake. The desired behavior from AUP planning here is to make the cake, and not worry about preserving the ability to make pasta or toast as neither input reward cares about it.

The Gaussian posterior fails to make the cake. The distribution over $\tout_{\text{cake}}$ here is a Gaussian with $\mu\approx 0.55$ and $\sigma\approx 0.71$. The posterior assigns substantial probability to $\tout_{\text{cake}}$ being negative, and AUP planning avoids making the cake to preserve the ability to optimize rewards that punish cakes.

Here each reward vector in $\tindep$ leads the agent to make cake. Since every method except Gaussian satisfies at least the weak version of Des.~3, all reward functions in the support of their posteriors lead the agent to make cake. Because of this AUP infers that it is not important to preserve the ability to make pasta or toast, and successfully makes cakes when applied to rewards sampled from these posteriors.

\subsection{Behavior balance}
To better understand robustness of the reward combination methods to imbalanced input rewards,
we analyze the performance of our methods by varying the extent of reward imbalance and the degree to which AUP seeks to preserve option value. 

The second input reward $\tin{2}$ is hand-specified to only reward toast, $\tin{2}_{\text{toast}}=1$, $\tin{2}_{\text{cake}}=\tin{2}_{\text{pasta}}=0$. The first reward $\tin{1}$ is hand-specified as well: 
$\tin{1}_{\text{toast}}=\tin{1}_{\text{pasta}}=0$. We vary $\tin{1}_{\text{cake}}$ between 0.01 and 0.7 to obtain many degrees of imbalance between toast and cake. So the input rewards disagree whether to make toast or cake, and agree that pasta is irrelevant. 
Hence the desirable action from AUP here would be to preserve the ability to choose between cake and toast, as either could be desirable.
In addition to varying $\tin{1}_{\text{cake}}$, we vary the weight of the AUP penalty~$\lambda$. 

\begin{figure}[ht]
  \centering
  \includegraphics[width=12cm]{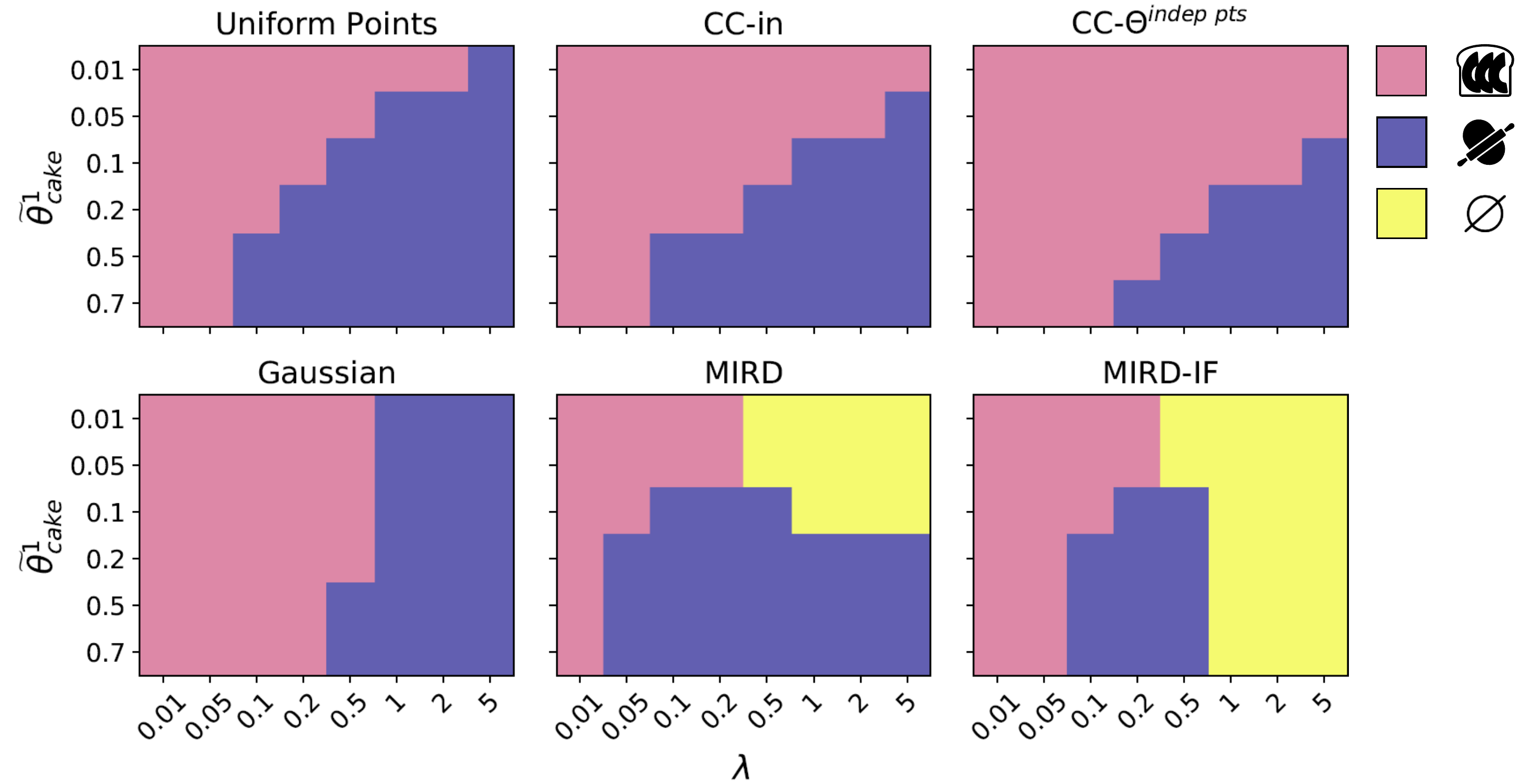}
  \caption{Most common outcomes of AUP applied to the six reward posteriors over 5 seeds. Pink corresponds to AUP failure for more than half of the seeds: $\tin{2}_{\text{toast}}$ overwhelms $\tin{1}_{\text{cake}}$ and the agent makes toast. Purple corresponds to AUP success (AUP makes dough) for more than half of the seeds. Yellow corresponds to the middle-ground outcome of doing nothing in more than half of the seeds. The hyperparameters $\beta$ of the $\Beta$ and the $\Dirichlet$ distributions are set to 0.5. 
  }
  \label{fig:aup_param_vs_reward}
\end{figure}

Additive always fails and makes toast as it has support on a single reward function that prioritizes toast. Our results for the other six reward combination methods are shown in Figure~\ref{fig:aup_param_vs_reward}.   We observe that MIRD and MIRD-IF are significantly more robust to imbalanced input rewards than any reward-space method. However, sometimes MIRD and MIRD-IF lead to AUP doing nothing instead of making dough. This is explained by the fact that MIRD and MIRD-IF use MCEIRL, which sometimes outputs negative rewards. For example, consider the case where in the process of generating a MIRD sample we generate 10 trajectories all of which demonstrate making toast. MCEIRL with these trajectories might output a positive reward on toast and a \textit{negative} reward on cake and dough (instead of positive reward for toast and zero for cake and dough), even if a zero-centered Gaussian prior is used. These negative rewards on dough sometimes lead AUP to avoid making it.




Furthermore, we observe that MIRD is more robust than MIRD-IF, and CC-in is more robust than CC-$\tindep$. This is because in this case support on rewards giving rise to behaviors arising from $\theta \in \tindep$ makes the reward distributions less balanced. In particular, denoting the value of $\tin{1}_{\text{cake}}$ we vary as $x$,
  \[
    \tindep_{\text{toast, cake}} = \left\{\begin{array}{lr}
        \theta_{\text{toast, cake}}=[0,0], \\
        \theta_{\text{toast, cake}}=[0,x], \\
        \theta_{\text{toast, cake}}=[1,0], \\
        \theta_{\text{toast, cake}}=[1,x] 
    \end{array}\right\}.
  \]
Here only the second reward vector incentivises the agent to make cake, while both the third and the fourth rewards incentivise the agent to make toast. This slight imbalance of $\tindep$ in favor of making toast lead to CC-$\tindep$ and MIRD-IF posteriors being more likely to cause AUP to make toast instead of preserving the ability to choose between cake and toast.

Finally, we observe that the Gaussian method is roughly equally robust to reward imbalance across the whole range of $\tin{1}_{\text{cake}}$. This is because unlike the other methods, the standard deviation of the Gaussian posterior does not depend $\tin{1}$ and $\tin{2}$, but only depends on hyperparameters $\sigma_1$ and $\sigma_2$, which are fixed to 1.

\end{document}